\documentclass[10pt,twocolumn,letterpaper]{article}

\usepackage[camera ready]{cvpr2025_stylekit/cvpr}      
\usepackage{cvpr}              

\usepackage{graphicx}
\usepackage{amsmath}
\usepackage{amssymb}
\usepackage{booktabs}

\usepackage{algorithm}
\usepackage{algpseudocode}
\usepackage{amsmath}
\usepackage{multirow}

\usepackage{bm}
\usepackage{rotating}

\usepackage{hyperref}
\hypersetup{pagebackref,breaklinks,colorlinks}

\usepackage[capitalize]{cleveref}
\crefname{section}{Sec.}{Secs.}
\Crefname{section}{Section}{Sections}
\Crefname{table}{Table}{Tables}
\crefname{table}{Tab.}{Tabs.}

\newcommand\rgt{\aftergroup\mathclose\aftergroup{\aftergroup}\right}

\newcommand{\topic}[2][0.5mm]{\vspace{#1}\noindent\textbf{#2}}

\def\paperID{13354} 
\def\confName{CVPR}
\def\confYear{2025}

\begin{document}

\title{MoViE: Mobile Diffusion for Video Editing}
\author{Adil Karjauv\thanks{Equal contribution} \qquad
Noor Fathima\footnotemark[1] \qquad
Ioannis Lelekas \qquad
Fatih Porikli \\
Amir Ghodrati \qquad
Amirhossein Habibian \\
{Qualcomm AI Research\thanks{Qualcomm AI Research is an initiative of Qualcomm Technologies, Inc}} \\
}


\newcommand\ah[1]{\textcolor{green}{AmirH: #1}} %
\newcommand\amirg[1]{\textcolor{red}{AmirG: #1}} %
\newcommand\adil[1]{\textcolor{orange}{Adil: #1}} %
\newcommand\noor[1]{\textcolor{blue}{Noor: #1}} %
\newcommand\ioannis[1]{\textcolor{cyan}{Ioannis: #1}} %

\newcommand{\unet}{\bm{\epsilon}}
\newcommand{\unetH}{\bm{\epsilon}_H}
\newcommand{\unetHin}{\bm{\epsilon}_H^{in}}
\newcommand{\unetHout}{\bm{\epsilon}_H^{out}}
\newcommand{\unetL}{\bm{\epsilon}_L}
\newcommand{\repin}[1]{\bm{r}_{#1}^{in}}
\newcommand{\repout}[1]{\bm{r}_{#1}^{out}}
\newcommand{\repoutpred}[1]{\hat{\bm{r}}_{#1}^{out}}
\newcommand{\adap}{\bm{\phi}_\theta}

\newcommand\methodname{MoViE\xspace}


\newcommand{\head}[1]{{\smallskip\noindent\textbf{#1}}}
\newcommand{\alert}[1]{{\color{red}{#1}}}
\newcommand{\sm}{\scriptsize}
\newcommand{\eq}[1]{(\ref{eq:#1})}

\newcommand{\Th}[1]{\textsc{#1}}
\newcommand{\mr}[2]{\multirow{#1}{*}{#2}}
\newcommand{\mc}[2]{\multicolumn{#1}{c}{#2}}
\newcommand{\tb}[1]{\textbf{#1}}
\newcommand{\ch}{\checkmark}

\newcommand{\red}[1]{{\color{red}{#1}}}
\newcommand{\blue}[1]{{\color{blue}{#1}}}
\newcommand{\green}[1]{\color{green}{#1}}

\newcommand{\citeme}[1]{\red{[XX]}}
\newcommand{\refme}[1]{\red{(XX)}}

\newcommand{\fig}[2][1]{\includegraphics[width=#1\linewidth]{fig/#2}}
\newcommand{\figh}[2][1]{\includegraphics[height=#1\linewidth]{fig/#2}}


\newcommand{\tran}{^\top}
\newcommand{\mtran}{^{-\top}}
\newcommand{\zcol}{\mathbf{0}}
\newcommand{\zrow}{\zcol\tran}

\newcommand{\ind}{\mathbbm{1}}
\newcommand{\expect}{\mathbb{E}}
\newcommand{\nat}{\mathbb{N}}
\newcommand{\zahl}{\mathbb{Z}}
\newcommand{\real}{\mathbb{R}}
\newcommand{\proj}{\mathbb{P}}
\newcommand{\prob}{\mathbf{Pr}}
\newcommand{\normal}{\mathcal{N}}

\newcommand{\mif}{\textrm{if}\ }
\newcommand{\other}{\textrm{otherwise}}
\newcommand{\minimize}{\textrm{minimize}\ }
\newcommand{\maximize}{\textrm{maximize}\ }
\newcommand{\st}{\textrm{subject\ to}\ }

\newcommand{\id}{\operatorname{id}}
\newcommand{\const}{\operatorname{const}}
\newcommand{\sgn}{\operatorname{sgn}}
\newcommand{\var}{\operatorname{Var}}
\newcommand{\mean}{\operatorname{mean}}
\newcommand{\trace}{\operatorname{tr}}
\newcommand{\diag}{\operatorname{diag}}
\newcommand{\vect}{\operatorname{vec}}
\newcommand{\cov}{\operatorname{cov}}
\newcommand{\sign}{\operatorname{sign}}
\newcommand{\prj}{\operatorname{proj}}

\newcommand{\softmax}{\operatorname{softmax}}
\newcommand{\clip}{\operatorname{clip}}

\newcommand{\defn}{\mathrel{:=}}
\newcommand{\peq}{\mathrel{+\!=}}
\newcommand{\meq}{\mathrel{-\!=}}

\newcommand{\floor}[1]{\left\lfloor{#1}\right\rfloor}
\newcommand{\ceil}[1]{\left\lceil{#1}\right\rceil}
\newcommand{\inner}[1]{\left\langle{#1}\right\rangle}
\newcommand{\norm}[1]{\left\|{#1}\right\|}
\newcommand{\abs}[1]{\left|{#1}\right|}
\newcommand{\frob}[1]{\norm{#1}_F}
\newcommand{\card}[1]{\left|{#1}\right|\xspace}
\newcommand{\divg}[2]{{#1\ ||\ #2}}
\newcommand{\diff}{\mathrm{d}}
\newcommand{\der}[3][]{\frac{d^{#1}#2}{d#3^{#1}}}
\newcommand{\pder}[3][]{\frac{\partial^{#1}{#2}}{\partial{#3^{#1}}}}
\newcommand{\ipder}[3][]{\partial^{#1}{#2}/\partial{#3^{#1}}}
\newcommand{\dder}[3]{\frac{\partial^2{#1}}{\partial{#2}\partial{#3}}}

\newcommand{\wb}[1]{\overline{#1}}
\newcommand{\wt}[1]{\widetilde{#1}}

\def\xssp{\hspace{1pt}}
\def\ssp{\hspace{3pt}}
\def\msp{\hspace{5pt}}
\def\lsp{\hspace{12pt}}

\newcommand{\cA}{\mathcal{A}}
\newcommand{\cB}{\mathcal{B}}
\newcommand{\cC}{\mathcal{C}}
\newcommand{\cD}{\mathcal{D}}
\newcommand{\cE}{\mathcal{E}}
\newcommand{\cF}{\mathcal{F}}
\newcommand{\cG}{\mathcal{G}}
\newcommand{\cH}{\mathcal{H}}
\newcommand{\cI}{\mathcal{I}}
\newcommand{\cJ}{\mathcal{J}}
\newcommand{\cK}{\mathcal{K}}
\newcommand{\cL}{\mathcal{L}}
\newcommand{\cM}{\mathcal{M}}
\newcommand{\cN}{\mathcal{N}}
\newcommand{\cO}{\mathcal{O}}
\newcommand{\cP}{\mathcal{P}}
\newcommand{\cQ}{\mathcal{Q}}
\newcommand{\cR}{\mathcal{R}}
\newcommand{\cS}{\mathcal{S}}
\newcommand{\cT}{\mathcal{T}}
\newcommand{\cU}{\mathcal{U}}
\newcommand{\cV}{\mathcal{V}}
\newcommand{\cW}{\mathcal{W}}
\newcommand{\cX}{\mathcal{X}}
\newcommand{\cY}{\mathcal{Y}}
\newcommand{\cZ}{\mathcal{Z}}

\newcommand{\vA}{\mathbf{A}}
\newcommand{\vB}{\mathbf{B}}
\newcommand{\vC}{\mathbf{C}}
\newcommand{\vD}{\mathbf{D}}
\newcommand{\vE}{\mathbf{E}}
\newcommand{\vF}{\mathbf{F}}
\newcommand{\vG}{\mathbf{G}}
\newcommand{\vH}{\mathbf{H}}
\newcommand{\vI}{\mathbf{I}}
\newcommand{\vJ}{\mathbf{J}}
\newcommand{\vK}{\mathbf{K}}
\newcommand{\vL}{\mathbf{L}}
\newcommand{\vM}{\mathbf{M}}
\newcommand{\vN}{\mathbf{N}}
\newcommand{\vO}{\mathbf{O}}
\newcommand{\vP}{\mathbf{P}}
\newcommand{\vQ}{\mathbf{Q}}
\newcommand{\vR}{\mathbf{R}}
\newcommand{\vS}{\mathbf{S}}
\newcommand{\vT}{\mathbf{T}}
\newcommand{\vU}{\mathbf{U}}
\newcommand{\vV}{\mathbf{V}}
\newcommand{\vW}{\mathbf{W}}
\newcommand{\vX}{\mathbf{X}}
\newcommand{\vY}{\mathbf{Y}}
\newcommand{\vZ}{\mathbf{Z}}

\newcommand{\va}{\mathbf{a}}
\newcommand{\vb}{\mathbf{b}}
\newcommand{\vc}{\mathbf{c}}
\newcommand{\vd}{\mathbf{d}}
\newcommand{\ve}{\mathbf{e}}
\newcommand{\vf}{\mathbf{f}}
\newcommand{\vg}{\mathbf{g}}
\newcommand{\vh}{\mathbf{h}}
\newcommand{\vi}{\mathbf{i}}
\newcommand{\vj}{\mathbf{j}}
\newcommand{\vk}{\mathbf{k}}
\newcommand{\vl}{\mathbf{l}}
\newcommand{\vm}{\mathbf{m}}
\newcommand{\vn}{\mathbf{n}}
\newcommand{\vo}{\mathbf{o}}
\newcommand{\vp}{\mathbf{p}}
\newcommand{\vq}{\mathbf{q}}
\newcommand{\vr}{\mathbf{r}}
\newcommand{\Vs}{\mathbf{s}}
\newcommand{\vt}{\mathbf{t}}
\newcommand{\vu}{\mathbf{u}}
\newcommand{\vv}{\mathbf{v}}
\newcommand{\vw}{\mathbf{w}}
\newcommand{\vx}{\mathbf{x}}
\newcommand{\vy}{\mathbf{y}}
\newcommand{\vz}{\mathbf{z}}

\newcommand{\vone}{\mathbf{1}}
\newcommand{\vzero}{\mathbf{0}}

\newcommand{\valpha}{{\boldsymbol{\alpha}}}
\newcommand{\vbeta}{{\boldsymbol{\beta}}}
\newcommand{\vgamma}{{\boldsymbol{\gamma}}}
\newcommand{\vdelta}{{\boldsymbol{\delta}}}
\newcommand{\vepsilon}{{\boldsymbol{\epsilon}}}
\newcommand{\vzeta}{{\boldsymbol{\zeta}}}
\newcommand{\veta}{{\boldsymbol{\eta}}}
\newcommand{\vtheta}{{\boldsymbol{\theta}}}
\newcommand{\viota}{{\boldsymbol{\iota}}}
\newcommand{\vkappa}{{\boldsymbol{\kappa}}}
\newcommand{\vlambda}{{\boldsymbol{\lambda}}}
\newcommand{\vmu}{{\boldsymbol{\mu}}}
\newcommand{\vnu}{{\boldsymbol{\nu}}}
\newcommand{\vxi}{{\boldsymbol{\xi}}}
\newcommand{\vomikron}{{\boldsymbol{\omikron}}}
\newcommand{\vpi}{{\boldsymbol{\pi}}}
\newcommand{\vrho}{{\boldsymbol{\rho}}}
\newcommand{\vsigma}{{\boldsymbol{\sigma}}}
\newcommand{\vtau}{{\boldsymbol{\tau}}}
\newcommand{\vupsilon}{{\boldsymbol{\upsilon}}}
\newcommand{\vphi}{{\boldsymbol{\phi}}}
\newcommand{\vchi}{{\boldsymbol{\chi}}}
\newcommand{\vpsi}{{\boldsymbol{\psi}}}
\newcommand{\vomega}{{\boldsymbol{\omega}}}

\newcommand{\rLambda}{\mathrm{\Lambda}}
\newcommand{\rSigma}{\mathrm{\Sigma}}

\newcommand{\vLambda}{\bm{\rLambda}}
\newcommand{\vSigma}{\bm{\rSigma}}


\makeatletter
\newcommand{\vast}[1]{\bBigg@{#1}}
\makeatother

\makeatletter
\newcommand*\bdot{\mathpalette\bdot@{.7}}
\newcommand*\bdot@[2]{\mathbin{\vcenter{\hbox{\scalebox{#2}{$\m@th#1\bullet$}}}}}
\makeatother

\makeatletter
\DeclareRobustCommand\onedot{\futurelet\@let@token\@onedot}
\def\@onedot{\ifx\@let@token.\else.\null\fi\xspace}

\def\eg{\emph{e.g}\onedot} \def\Eg{\emph{E.g}\onedot}
\def\ie{\emph{i.e}\onedot} \def\Ie{\emph{I.e}\onedot}
\def\cf{\emph{cf}\onedot} \def\Cf{\emph{Cf}\onedot}
\def\etc{\emph{etc}\onedot} \def\vs{\emph{vs}\onedot}
\def\wrt{w.r.t\onedot} \def\dof{d.o.f\onedot} \def\aka{a.k.a\onedot}
\def\etal{\emph{et al}\onedot}
\makeatother


\twocolumn[{%
\renewcommand\twocolumn[1][]{#1}%
    \maketitle

\begin{center}
    \centering
    \captionsetup{type=figure, skip=0pt}
    \vspace{-0.8cm}
\includegraphics[width=1.8\columnwidth]{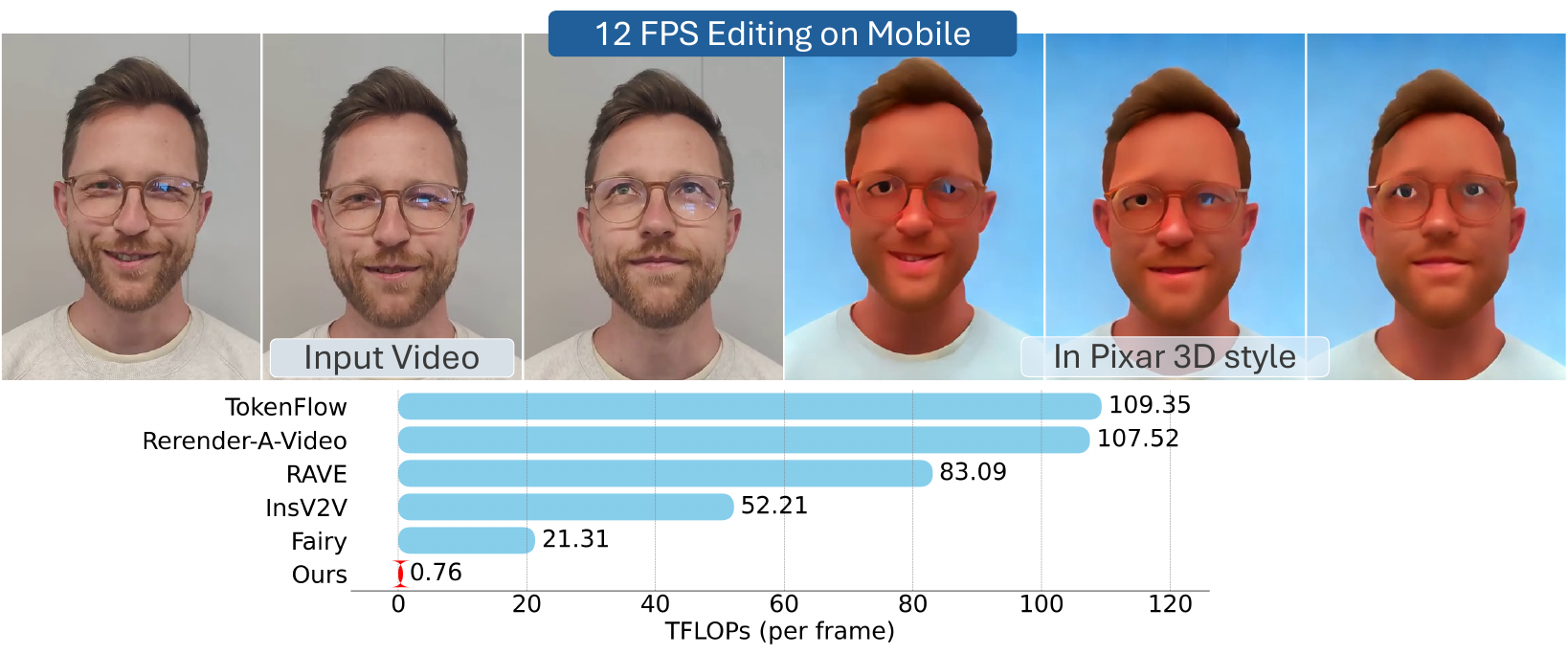}
    \captionof{figure}{\methodname is a fast video editing model, capable of generating $12$ frames per second on a mobile phone. It requires significantly fewer floating point operations (FLOPs) to edit a single video frame, making it computationally more efficient than competing methods.
    }
\end{center}%
}
]
{\renewcommand{\thefootnote}{\fnsymbol{footnote}}\stepcounter{footnote}%
  \footnotetext{Equal Contribution}}
{\renewcommand{\thefootnote}{\fnsymbol{footnote}}\stepcounter{footnote}%
\footnotetext{Qualcomm AI Research is initiative of Qualcomm Technologies, Inc. Snapdragon and Qualcomm branded products are products of Qualcomm Technologies, Inc. and/or its subsidiaries.}}

\begin{abstract}
Recent progress in diffusion-based video editing has shown remarkable potential for practical applications. 
However, these methods remain prohibitively expensive and challenging to deploy on mobile devices. In this study, we introduce a series of optimizations that render mobile video editing feasible. Building upon the existing image editing model, we first optimize its architecture and incorporate a lightweight autoencoder. Subsequently, we extend classifier-free guidance distillation to multiple modalities, resulting in a threefold on-device speedup. 
Finally, we reduce the number of sampling steps to one by introducing a novel adversarial distillation scheme which preserves the controllability of the editing process. 
Collectively, these optimizations enable video editing at $12$ frames per second on mobile devices, while maintaining high quality. Our results are available at \url{https://qualcomm-ai-research.github.io/mobile-video-editing/}.
\end{abstract}

\section{Introduction}
Diffusion-based generative models have shown impressive quality in image editing~\cite{hertz2022prompt, tumanyan2023plug, ip2p, parmar2023zero}. Building on these advancements, zero-shot video-based models extend these capabilities to the temporal dimension for video editing~\cite{wu2023tune, qi2023fatezero, geyer2023tokenflow, ceylan2023pix2video, wu2024fairy, liu2024video}. This is typically achieved by developing techniques to propagate information between frames ensuring that changes made in one frame are coherently reflected in subsequent frames. While these techniques achieve remarkable results in style transfer and attribute editing, they are computationally expensive, making them prohibitive to run on constrained edge devices,~\eg mobile phones. 

While significant efforts have been made to reduce the computational cost of image diffusion models for on-device generation~\cite{li2024snapfusion,zhao2023mobilediffusion, choi2023squeezing, castells2024edgefusion}, there has been little effort towards developing video editing models on edge, mainly due to the aforementioned high computational and memory costs. Inversion-based models~\cite{qi2023fatezero, geyer2023tokenflow, kara2024rave} that extract source video features using DDIM inversion incur high computational and memory costs during inference. On the other hand, methods that propagate features from one or more frames to the other, a process known as cross-frame attention~\cite{wu2023tune}, also remain expensive to run on-device, as each frame must attend to multiple frames. For example, editing a 120-frame video at a resolution of $512\times384$ with Fairy~\cite{wu2024fairy}, a fast video-to-video (V2V) model, takes $13.8$ seconds on $8\times$A100 GPUs. This is even higher for TokenFlow~\cite{geyer2023tokenflow}, an inversion-based V2V model, which takes $744$ seconds~\cite{wu2024fairy}.

In this work, we introduce the first on-device video editing model capable of generating $12$ frames per second on Xiaomi-14 Pro. Xiaomi-14 Pro uses a Snapdragon\textsuperscript{\textregistered} 8 Gen. 3 Mobile Platform with a Hexagon\textsuperscript{\tiny{TM}} processor. 
To achieve this, we begin with a conventional video editing pipeline. Broadly speaking, diffusion editing pipelines~\cite{wu2023tune, wu2024fairy} first encode source frames into the latent space using a Variational AutoEncoder (VAE)~\cite{pinheiro2021variational}. A denoising UNet model then iteratively denoises latents to generate edited latents, which are then decoded by the VAE decoder to produce the edited frames. In addition, many diffusion models use the classifier-free guidance technique to balance quality and diversity. Specifically, editing models utilize two guidance scales — image and text — which can be adjusted to trade off how strongly the generated image correspond with the source image and the edit instructions, respectively. 

In this work, we identify the factors causing computational bottlenecks within an editing pipeline. 
First, we observed that the architectures of both the denoising UNet and the VAE encoder/decoder are major bottlenecks in the editing pipeline. 
Second, while classifier-free guidance technique enhances the model’s controllability, it increases the computational cost three times as we need 3 forward passes per diffusion step to obtain the output.
Lastly, the intrinsic design of diffusion models necessitates iterative denoising to edit frames. Each of denoising steps involves complex computations, making the generation process expensive.
To address the aforementioned challenges, we propose a series of contributions:
\begin{itemize}
    \item We introduce the first on-device video-to-video diffusion model, capable of editing a $120$-frame video at a resolution of $512\times384$ on a mobile device in $9.6$ seconds, achieving a generation rate of $12$ fps. 
    \item We propose a method to jointly distill multiple guidance scales in the editing models, reducing the Number of Forward Evaluations (NFE) per diffusion step by $3$ times.
     \item We present an adjusted adversarial distillation recipe for reducing number of diffusion steps, ensuring the model’s controllability is maintained post-distillation while reducing NFE of denoiser by $10$ times.
\end{itemize}

Additionally, unlike some methods~\cite{wu2023tune} that can generate only a limited number of frames due to memory constraints, our model can operate on long videos. This is because our model generate video frames independent of the video length, and its complexity increases linearly with the number of frames.
\section{Related work}
\topic{Video generation.} 
Recent methods can be categorised into two groups. The first group enhances image models with temporal components, i.e. 3D convolutions, spatio-temporal transformers \cite{ho2022imagen, ho2022video, hong2022cogvideo, singer2022make, zhou2022magicvideo}, for modeling temporal dimension at the expense of the inflated computational footprint, along with the need for extensive training on large and diverse collections of data. 
In the effort to overcome these hurdles, zero-shot models \cite{kahatapitiya2025object}, rely on training-free techniques for boosting temporal consistency \cite{geyer2023tokenflow, khachatryan2023text2video, liu2024video, qi2023fatezero, zhang2023controlvideo}. Common methods are, utilization of structural conditioning from source video \cite{zhang2023controlvideo, chen2023control}, injection of diffusion inversion features into the generation process \cite{geyer2023tokenflow, qi2023fatezero} and employing cross-frame attention using a set of key frames \cite{qi2023fatezero, zhang2023controlvideo, wu2023tune, esser2023structure}. Our method aligns with this direction but is computationally much lighter, making it suitable for low-end edge devices.

\topic{Optimized diffusion.}
Optimizing diffusion mainly entails the reduction of required steps for traversing the probability path from pure noise to the denoised sample, as well as reducing the cost of each step. In line with the former are the utilization of higher order solvers \cite{zhang2022fast, lu2022dpm, lu2022dpmpp}, the straightening of underlying ODE trajectories or direct mapping to data via Rectified Flows \cite{liu2022flow, liu2022rectified, zhu2025slimflow} and consistency models \cite{lu2024simplifying, song2023consistency, song2023improved}, respectively and the employment of progressive-step distillation \cite{li2024snapfusion, salimans2022progressive, meng2023distillation} or adversarial training \cite{wang2022diffusion, sauer2024fast, sauer2025adversarial, zhang2024sf} for fewer or even single step diffusion. On the other hand, quantization \cite{he2024ptqd, pandey2023softmax, shang2023post} and pruning \cite{li2024snapfusion, choi2023squeezing} have been used, as well as extensive research conducted on simplifying the denoiser itself \cite{dockhorn2023distilling, kim2023bk, habibian2024clockwork}.
Our work incorporates both architectural optimization and a reduction in the number of steps. While extending \cite{li2024snapfusion} with multimodal distillation, we manage to address a limitation of \cite{sauer2024fast, zhang2024sf}, by restoring control over text and image guidance, the level of adherance to text and image conditioning respectively.

\topic{On-device generation.}
The abundance of edge devices, along with reduced cost and more privacy-secure inference on edge comparing to cloud-based approaches, have led to recent works targeting mobile devices and their NPUs \cite{castells2024edgefusion, chen2023speed, choi2023squeezing, li2024snapfusion, zhao2023mobilediffusion}. 
While there has been progress in the video domain with fast zero-shot video editing models~\cite{zhang2024fastvideoedit, kara2024rave, wu2024fairy}, there have been no attempts to implement  video editing models on device due to their high computational and memory requirements.
Our method pushes the boundaries by unlocking on-device zero-shot video editing.

\section{\methodname}
Our goal is to develop an on-device video-editing model that optimizes trade-off between efficiency and quality. To achieve this, we follow four stages: (1) selecting a base model, (2) introducing a mobile-friendly denoiser called Mobile-Pix2Pix, (3) reducing the cost of classifier-free guidance through multimodal guidance distillation, and (4) reducing number of steps using adversarial step distillation.

\subsection{Base model}
Zero-shot video editing using diffusion models have achieved impressive results. The common approach among such models is to employ an off-the-shelf image editing model and replace self-attention layers with cross-frame attention to ensure temporally consistent and coherent frame generation.
In specific, each self-attention layer receives the latent from previous layer and linearly projects it into query, key and value $Q,K,V$ to produce the output by Self-Attn($Q,K,V$) = $Softmax( QK^T/ \sqrt{d} )V$. Cross-frame attention extends the idea by concatenating keys and values from one or more other frames, commonly called as anchors, resulting in CrossFrame-Attn($Q,K^\prime,V^\prime$) = $Softmax( Q[K_1;,...;K_t]^T/\sqrt{d})[V_1;...;V_t]$.
For example, Rerender-A-Video~\cite{yang2023rerender} uses Stable Diffusion~\cite{rombach2022high} as image editing model and uses first and previous frames as anchors to attend to the current frame. Fairy~\cite{wu2024fairy} uses InstructPix2Pix~\cite{ip2p} as an instruction-based image editing model and uniformly selects 3 frames with equal intervals as anchors.
Following the same direction, we use InstructPix2Pix as the base image model. InstructPix2Pix processes a source image and a text instruction for the desired edit. The text is encoded using a text encoder (like CLIP~\cite{radford2021learning}), and the source image is converted into latent representations via a Variational Auto-Encoder (VAE). A conditional U-Net model then refines these latents iteratively for a fixed number of sampling steps to produce the edited image, which is finally decoded back to pixel space using VAE decoder. Frames are generated in 10 diffusion steps, with the middle frame used as the sole anchor.
In the subsequent sections, we detail our optimization strategies for the base model. 

\subsection{Mobile-Pix2Pix}
InstructPix2Pix pipeline is computationally intensive. For instance, denoising a single step of batch 1 with UNet requires approximately $600$ GFLOPs. For 10 steps and three forward passes needed for classifier-free guidance, this totals to $\sim18.05$ TFLOPs. Additionally, encoding and decoding an image of resolution $480 \times 480$ using VAE costs $3.2$ TFLOPs. We implement two changes to optimize InstructPix2Pix pipeline for video editing on mobile devices. First, similar to~\cite{li2024snapfusion}, we remove expensive self-attention and cross-attention layers at the highest resolutions, in both encoder and decoder part of the UNet. Removing them leads to $12\%$ reduction in FLOPs.

\begin{algorithm}[h!]
\caption{Multimodal Guidance Distillation}
\label{alg:mult_guid_dist}
\begin{algorithmic}[1]
\Require Teacher $\epsilon$, Student ${M}_{\theta}$, dataset $D$, loss weighting term $\lambda$, learning rate $\gamma$
\While{not converged}
    \State $c_I, c_T, x \sim D$ \Comment{Sample input image, prompt, edited image}
    \State $t \sim U[0, 1000]$ \Comment{Sample time}
    \State $s_I \sim U[1, 3]$ \Comment{Sample image guidance scale}
    \State $s_T \sim U[2, 14]$ \Comment{Sample text guidance scale}
    \State $\epsilon_n \sim N(0, I)$ \Comment{Sample noise}
    \State $x_t = \alpha_t x + \sigma_t \epsilon_n$ \Comment{Add noise to data}
    \State Obtain $\tilde{\epsilon}$ from Eq. 1  \Comment{CFG}
    \State $L = \lambda \left\| {M}_{\theta}(x_t, c_I, c_T, s_I, s_T) - \tilde{\epsilon} \right\|_2^2$ \Comment{Loss}
    \State $\theta \leftarrow \theta - \gamma \nabla_{\theta} L$ \Comment{Optimization}
\EndWhile
\end{algorithmic}
\end{algorithm}


Second, we reduce the latency induced by encoding and decoding latents using VAE by utilizing the Tiny Autoencoder for Stable Diffusion (TAESD) \cite{taesd}. TAESD is a deterministic and lightweight autoencoder composed of a series of residual blocks. Its encoder is a distilled version of the VAE encoder, whereas the decoder has been trained as a stand-alone GAN, optimized with adversarial and reconstruction losses.
Moreover, its shared latent space with VAE enables hybrid autoencoding, with VAE decoding TAESD's latents and vice versa. Incorporating TAESD in our pipeline leads to $92.6\%$ FLOPs savings compared to original InstructPix2Pix VAE with a modest drop in quality (see \cref{sec:ablations}). 

\subsection{Multimodal Guidance Distillation}
\label{sec:multimodal_guidance_distillation}
Classifier-free guidance (CFG) is a technique to improve the quality of text-to-image generative models. It is achieved by a linear combination of estimates from text-conditional and unconditional models, weighted by a so-called guidance scale. While this technique allows for trade-off between fidelity and diversity, it incurs computational costs due to two forward passes in each diffusion step, with and without text condition. To reduce this cost, ~\cite{meng2023distillation} distill classifier-free guidance into a model that requires only one forward pass per diffusion step in single modality, i.e. text. However, several works have utilized \textit{multi-modal} CFG for image-to-image editing~\cite{ip2p}, editing 3D scenes with text~\cite{haque2023nerf}, novel view synthesis~\cite{liu2023zero}, and video generation~\cite{kondratyuk2023videopoet}. In this work, we extend the concept of CFG distillation to multiple modalities, specifically text and image conditioning, to further reduce the cost of InstructPix2Pix pipeline.

\begin{figure}[tb]
\centering
\includegraphics[width=\columnwidth]{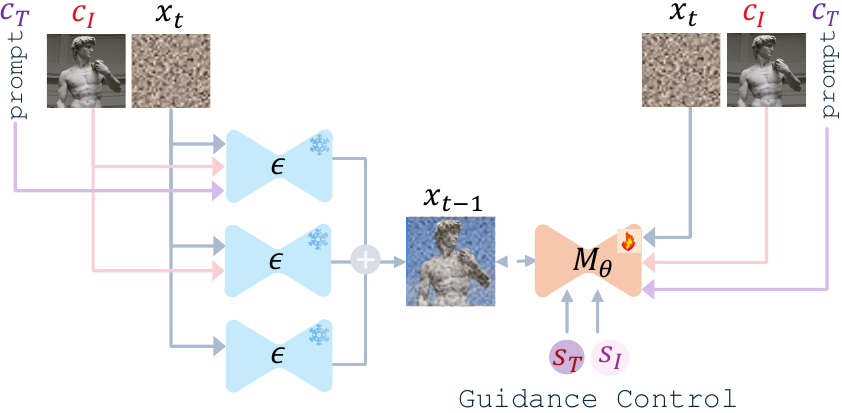}
\caption{Multimodal Guidance Distillation Overview: Standard classifier-free guidance inference pipeline (left) with two input conditionings (image and text) requires three inference runs per diffusion step. Our distilled pipeline (right) incorporates guidance scales $s_I$ and $s_T$ into UNet and only performs one inference run.
}
\label{fig:multimodal_guidance_distill}
\end{figure}

InstructPix2Pix model accepts source image condition $c_I$ and text condition $c_T$ as input. During inference, the model performs three separate forward passes with different conditionings, that are subsequently combined to generate the final output~\cite{ip2p}:

\begin{equation}
\setlength{\abovedisplayskip}{5pt}
\setlength{\belowdisplayskip}{5pt}
\begin{aligned}
\tilde{\epsilon}(x_t, c_I, c_T) &= \epsilon(x_t, \varnothing, \varnothing) \\
&\quad + s_I \left( \epsilon(x_t, c_I, \varnothing) - \epsilon(x_t, \varnothing, \varnothing) \right) \\
&\quad + s_T \left( \epsilon(x_t, c_I, c_T) - \epsilon(x_t, c_I, \varnothing) \right)
\end{aligned}
\label{eq:1}
\end{equation}
where $\epsilon(.)$ is a denoiser function and $s_I$ and $s_T$ are image and text guidance scales respectively. $s_I$ and $s_T$ are control signals that can be adjusted to trade off the edited image fidelity to the source image and the edit instruction, respectively. 
As shown in~\cref{fig:multimodal_guidance_distill}, we distill $\epsilon(.)$ in~\cref{eq:1} to a single forward pass using the loss function:
\begingroup
\setlength{\abovedisplayskip}{5pt} 
\setlength{\belowdisplayskip}{5pt}
\[
\mathbb{E}_{c_I, c_T \sim p_{\text{data}}}\left[\left\| {M}_{\theta}(x_t, c_I, c_T, s_I, s_T) - \tilde{\epsilon}(x_t, c_I, c_T) \right\|_{2}^{2} \right],
\]
\endgroup
where ${M}_{\theta}(.)$, a student model, receives guidance scales $s_I$ and $s_T$ along with the conditions. This is done first by projecting these scales to embeddings using sinusoidal timestep embeddings~\cite{ho2020denoising}. Each ResNet block is then modified to accept these embeddings similarly to timestep embedding, which are further processed through a linear layer and nonlinearity before being added to the latent representation. Note that both image and text guidance scales are distilled at once through our distillation procedure. For more details see~\cref{alg:mult_guid_dist}. As a result, the distilled model ${M}_{\theta}(.)$ allows for a single forward pass in each diffusion step without compromising the quality or control (See~\cref{fig:mobile_pix2pix_mult_guid_dist_clip_plot}).

\subsection{Adversarial Step Distillation}
\begin{figure}[tb]
\centering
\includegraphics[width=1.0\columnwidth]{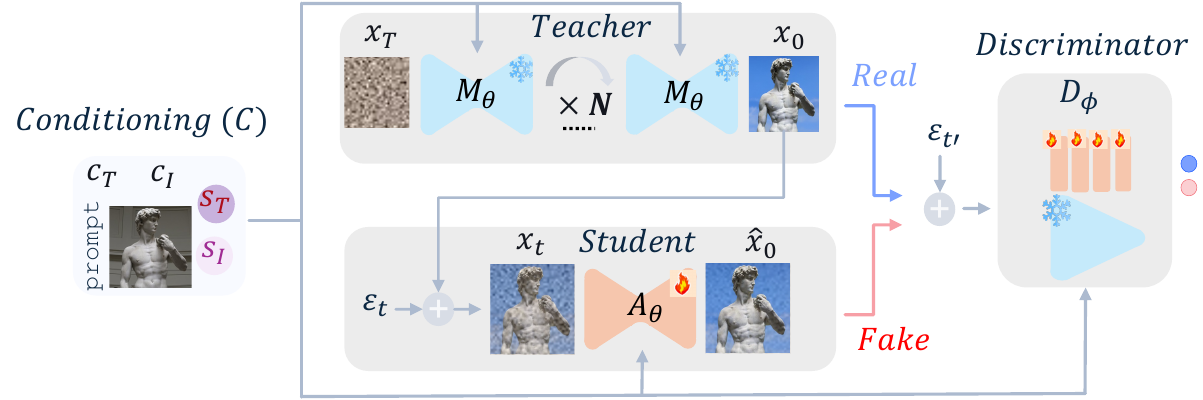}
\caption{Adversarial Distillation: We distill a multi-step teacher into a single step student using adversarial losses. Unlike existing adversarial distillation approaches \cite{sauer2024fast, zhang2024sf} that forego guidance flexibility for faster sampling, we preserve guidance strength property during adversarial training by providing the synthetic latent $x_t$ from teacher's denoising process and conditioning the student on the corresponding guidance scales. 
}
\label{fig:adversarial_distillation}
\end{figure}

Having reduced the computational footprint of each step, in this section, we seek to distill the multi-step model obtained in \cref{sec:multimodal_guidance_distillation} into a single-step model to establish an efficient on-device editing pipeline. 
Improving sampling efficiency of diffusion models has been the focus of several works~\cite{li2024snapfusion, salimans2022progressive, lu2024simplifying}. Recently, adversarial distillation diffusion models that require a single denoising step have shown impressive results for text-to-image (T2I) and image-to-image (I2I) editing applications~\cite{sauer2024fast, sauer2025adversarial, zhang2024sf}. However, the classifier-free guidance property is usually ignored during training~\cite{sauer2024fast, zhang2024sf}. For example LADD~\cite{sauer2024fast} foregoes the flexibility of text and image guidance in favor of increased sampling speed. This oversight of guidance scales leads to models that lack control over edit strength during inference, making them less practical for use. In this work, we propose an adversarial training algorithm, built upon LADD~\cite{sauer2024fast}, that distills a multi-step teacher into a single-step student while preserving its controllability. 
Our adversarial distillation setup is outlined in~\cref{fig:adversarial_distillation}. Student $A_{\theta}$ is initialized from pre-trained teacher $M_\theta$ of the previous stage. Discriminator $D_\phi$ consists of a frozen feature extractor, initialized from the encoder arm of the teacher and a set of trainable spatial heads applied to each layer of feature extractor. Detailed architecture of spatial heads can be found in the~\cref{sec:appendix:training_details}.
During adversarial training, $real$ edited sample $x_0$ is obtained from the multi-step teacher. Student receives a noisy sample $x_t$ generated by applying forward diffusion process to $x_0$ using student noise schedule. To retain controllability, the corresponding guidance scales $s_I, s_T$ are also provided to the student. The student then generates the $fake$ samples $\hat{x_0}$ which is then evaluated by the discriminator. Similar to \cite{sauer2024fast}, we first add noise to $x_0$ and $\hat{x_0}$ before feeding them to the discriminator. For more details see~\cref{alg:adv_dist}. Discriminator distinguishes between $real$ and $fake$ samples, providing feedback to the student model to improve its ability to generate high-quality, controlled edits.

\begin{algorithm}[H]
\caption{Adversarial Distillation}
\label{alg:adv_dist}
\begin{algorithmic}[1]
\Require Teacher model $M$, Student model ${A}_{\theta}$, Discriminator $D_\phi$, loss weighting term $\lambda$, learning rate $\gamma$ and $\gamma_d$ for $A$ and $D$, timesteps $T_g$ for $A$, preconditioning functions $c_{in}$ and $c_{in}'$~\cite{karras2022elucidating}, $R1$ is gradient penalty \cite{mescheder2018trainingmethodsgansactually}
\While{not converged}
    \State $c_I, c_T \sim D$, $s_I \sim U[1, 3]$, $s_T \sim U[2, 14]$
    \State $C = [c_I, c_T, s_I, s_T]$
    \State $\epsilon \sim N(0, I);  x_0 = M(\epsilon, C)$ \Comment{Real sample from teacher's denoising process}
    \State $t \sim U[0, 8], \epsilon \sim N(0, I)$
    \State $x_t = c_{in}(t) * (\alpha_t x_0 + \sigma_t \epsilon)$ \Comment{Add noise to real}
    \State $\hat x_0 = {A}_{\theta}(x_t, C, t)$ \Comment{Fake (Student) sample}
    \State $t' \sim U[0, 1000], \epsilon' \sim N(0, I)$
    \State $x_{t'} = c_{in}' * (\alpha_{t'} x_0 + \sigma_{t'} \epsilon')$ \Comment{Add noise to real}
    \State $\hat x_{t'} = c_{in}' * (\alpha_{t'} \hat x_0 + \sigma_{t'} \epsilon')$  \Comment{Add noise to fake}
    \State $L = \mathbb{E}_{t', x_0}[max(0, 1+D_\phi(x_{t'}, C)) + \lambda_{r1} R1] + \mathbb{E}_{t, t', x_0}[max(0, 1-D_\phi(\hat x_{t'}, C))]$ 
    \State $\phi \leftarrow \phi - \gamma_d \nabla_{\phi} L$ \Comment{Update Disc.}
    \State Repeat Step-8 and 10
    \State $L = \lambda_{mse} \left\| \hat x_0 -  x_0 \right\|_2^2 + \lambda_{gen} \mathbb{E}_{t, t', x_0} [  D_\phi(\hat x_t', C)]$
    \State $\theta \leftarrow \theta - \gamma \nabla_{\theta} L$ \Comment{Update Student}
\EndWhile
\end{algorithmic}
\end{algorithm}

\section{Experiments}

\subsection{Experimental setup}
\topic{Datasets and metrics.} 
In our work, we utilize the InstructPix2Pix dataset \cite{ip2p} for finetuning and distillation. The dataset has around $300k$ samples, each sample consisting of (source image, edit instruction, and edited image) triplet. For evaluation and ablations, we use the validation set consisting of around $5k$ samples. Following \cite{ip2p}, we utilize CLIP-Image similarity and directional CLIP similarity to estimate fidelity of edited image to the input image and edit prompt respectively. Higher values for these metrics indicate superior performance.
Following \cite{insv2v, fdd}, for comparison with the state-of-the-art, we employ the Text-Guided Video Editing (TGVE) benchmark \cite{wu2023cvpr}. The dataset comprises of $76$ videos where each video includes 4 prompts on editing style, background, object, and multiple attributes. We measure PickScore \cite{kirstain2023pickapicopendatasetuser} and CLIPFrame\cite{hessel2022clipscorereferencefreeevaluationmetric} to evaluate the quality and consistency of the edits respectively. 
To measure number of floating point operations (FLOPs), we use DeepSpeed~\cite[v0.15.2]{deepspeed}. Latencies are measured on a single NVIDIA\textsuperscript{\textregistered} A100 GPU and on a Xiaomi-14 Pro for GPU and Phone respectively. Xiaomi-14 Pro uses a Qualcomm Snapdragon\textsuperscript{\textregistered} 8 Gen. 3 Mobile Platform with a Hexagon\textsuperscript{\tiny{TM}} processor. Reported latencies correspond to a single-frame denoising.

\begin{figure}[tb]
\centering
\includegraphics[ width=\columnwidth]{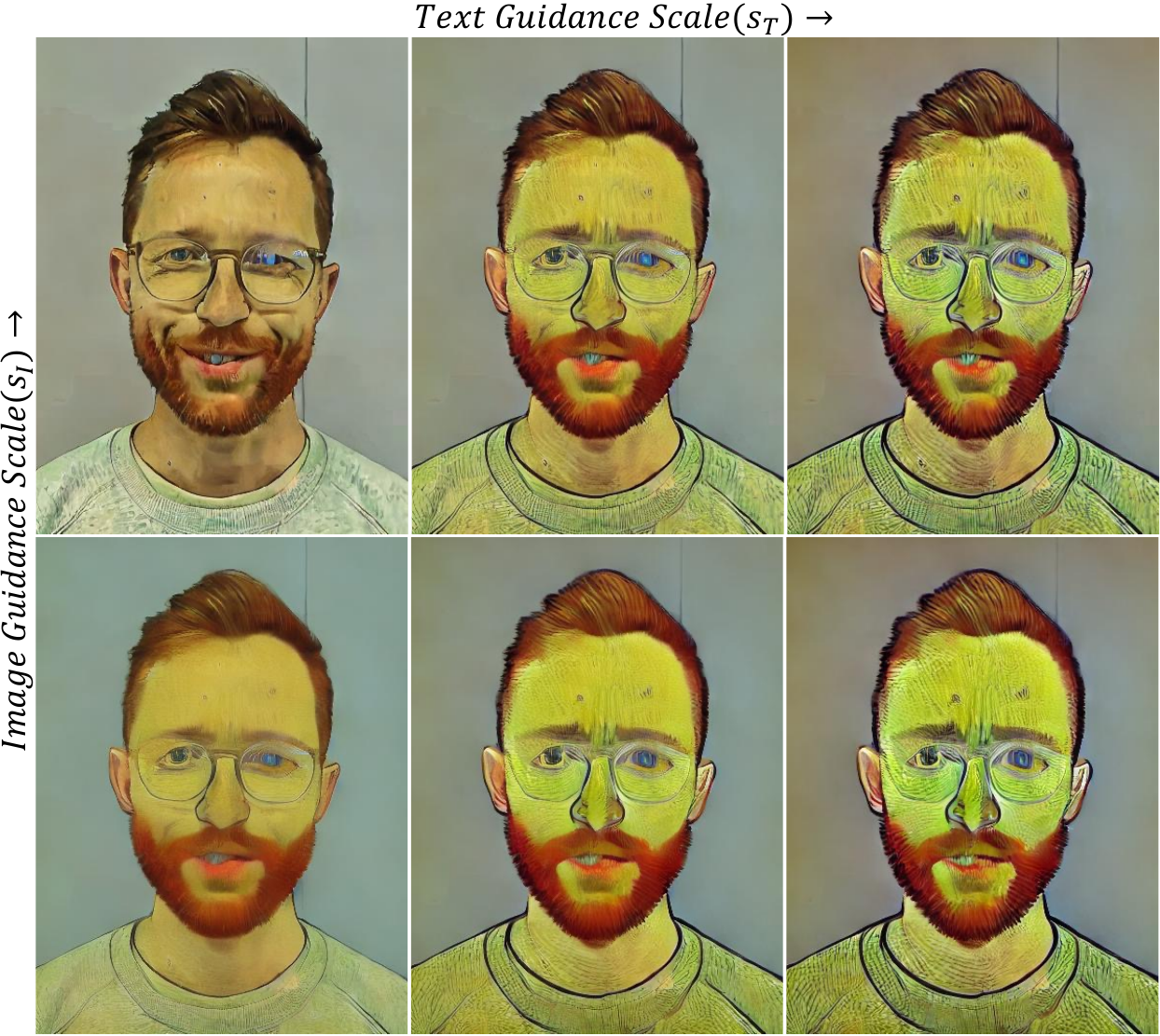}
\caption{MoViE at text guidance $[4.0,8.0,12.0]$ and image guidance $[1.25, 1.75]$. Our adversarial training maintains guidance scales, allowing us to control edit strength during inference. (Prompt: In Van Gogh Style)}
\label{fig:gan_flexibility}
\end{figure}

\topic{Implementation details.}
\label{sec:implementation_details}
To obtain an efficient and mobile-friendly model of InstructPix2Pix, we make sequential architectural and inference optimizations as follows. 
\textbf{Stage 1}: We begin with the text-to-image Stable Diffusion (SD) UNet V-1.5 UNet. Inspired by \cite{li2024snapfusion}, we remove the costly self-attention and cross-attention layers at the highest resolution and finetune the model. 
\textbf{Stage 2}: Similar to \cite{ip2p}, we add image conditioning layers and transform the T2I model to an I2I editing model and train on InstructPix2Pix dataset for 20k iterations, resulting in Mobile-Pix2Pix. \textbf{Stage 3}: We distill multimodal guidance scales as described in~\cref{alg:mult_guid_dist}. We use 5k iteration for training where the teacher is a model from the previous stage. \textbf{Stage 4}: For guidance-preserving adversarial distillation, we first convert the model obtained from the previous stage to $v-prediction$ as we find that it helps to reduce noisy artifacts during adversarial training. We then distill a multi-step teacher model into a single-step student using a weighted combination of $MSE$ and $GAN$ loss for 20k iterations. The detailed training steps are shown in \cref{alg:adv_dist}, and further details can be found in the Appendix. 

\subsection{Results}
\begin{figure}[tb]
\centering
\includegraphics[width=\columnwidth]{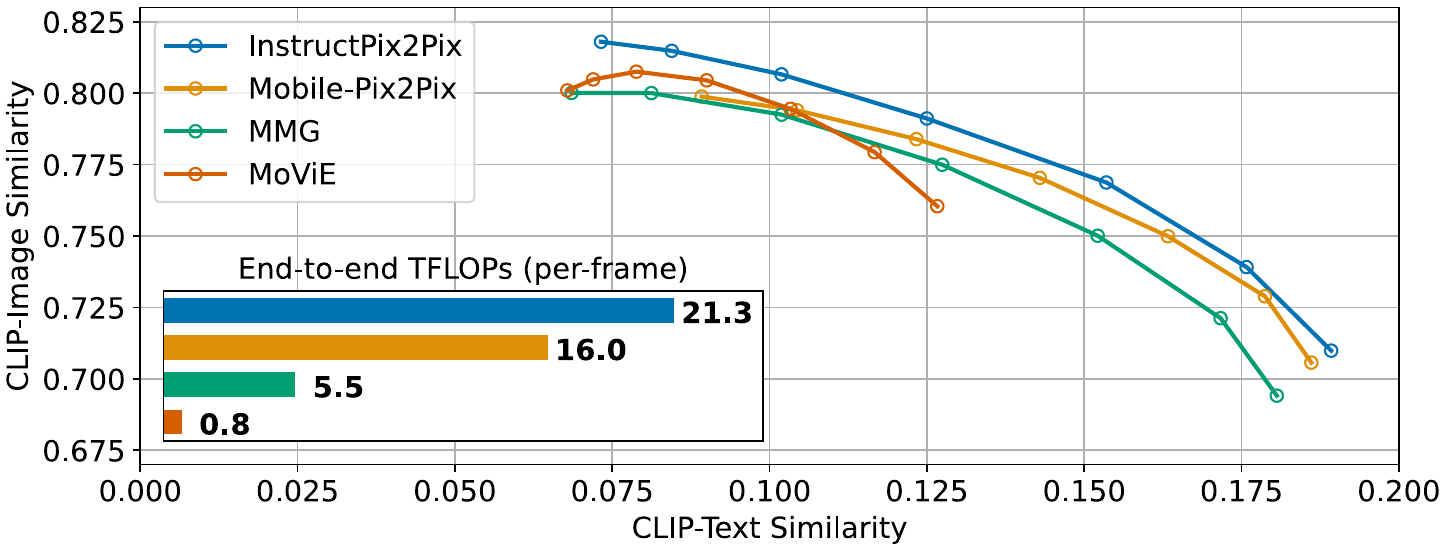}
\caption{CLIP metrics for InstructPix2Pix, Mobile-Pix2pix, Multi-modal Guidance (MMG) Mobile-Pix2pix and MoViE. As shown in the graphs, proposed optimizations improve the efficiency greatly with minimum quality drop.}
\label{fig:mobile_pix2pix_mult_guid_dist_clip_plot}
\end{figure}

\begin{table*}[t!]
\centering
\resizebox{2\columnwidth}{!}{
\def\arraystretch{1.}%
\begin{tabular}{lcccccccc}
\toprule
   \textbf{Method} & \textbf{Steps} & \textbf{PickScore}$\uparrow$ & \textbf{CLIPFrame}$\uparrow$ & \textbf{TFLOPs (per frame)} & \textbf{Latency} \textcolor{gray}{\small {(GPU)}} & \textbf{Latency} \textcolor{gray}{\small {(Phone)}}  \\

   \midrule
    Fairy                   & 10 & 19.80 & 0.933 & - & - & -\\   
    TokenFlow               & 50 & 20.49 & 0.940 & 109.35 & 2.45s & - \\
    Rerender-A-Video          & 20 & 19.58 & 0.909 & 107.52 & 2.13s & - \\
    ControlVideo            & 50 & 20.06 & 0.930 & 89.49 & 5.63s & - \\
    InsV2V                  & 20 & 20.76 & 0.911 & 52.21 & 2.70s & - \\
    RAVE                  & 50 & 20.35 & 0.932 & 83.09 & 4.31s & - \\
    EVE                  & - & 20.76 & 0.922 & - & - & - \\
    \midrule
    Base Model                                              & 10 & 20.34 & 0.943 & 21.31 & 1.37s & 7s\\
     \;           + Mobile-Pix2Pix        & 10 & 19.43 & 0.922  & 16.10 & 1.06s & 1.9s\\
     \;\;\;\;     + Multi-Guidance Dist.        & 10 & 19.60 & 0.919 & 5.50 & 0.82s & 0.6s \\
     \;\;\;\;\;   + Adversarial Distillation  & 1 & 19.40 & 0.913 & 0.76 & 0.11s & 0.08s\\
\bottomrule
\end{tabular}
}
\caption{End-to-end FLOPs and latency of video editing models on $480\times480$ resolution on \textbf{TGVE} benchmark, normalized per frame. On-device latencies reported on $512\times384$ frames. PickScore and CLIPFrame for competing methods (except RAVE) are from InsV2V~\cite{insv2v}.}
\label{tab:sota_v2v_tgve}
\vspace{-0.7em}
\end{table*}

\topic{MoViE.} We compare the impact of different optimization stages on editing quality in terms of directional CLIP similarity and CLIP-image similarity. As shown in~\cref{fig:mobile_pix2pix_mult_guid_dist_clip_plot}, our first optimization stage, Mobile-Pix2Pix, achieves comparable quality as the original model while being $24.4\%$ cheaper in terms of FLOPS and $73\%$ faster on-device (see~\cref{tab:sota_v2v_tgve}). With multimodal guidance distillation, a single forward pass is now required, thus our pipeline becomes $66\%$ cheaper computation-wise compared to previous stage, still at the cost of negligible quality drop. 
Finally, our adversarial distillation technique further improves UNet latency by a factor of $10$ and end-to-end latency by a factor of $7.5$, enabling on-device editing at $12$ frames per second. In the last stage, we observe a slight decline in quantitative metrics; however, the substantial gains in efficiency justify the trade-off, as evidenced by~\cref{tab:sota_v2v_tgve} and the qualitative results in~\cref{fig:qual_results_faces}. Finally, in ~\cref{fig:gan_flexibility} we show that our adversarial distillation technique retains the flexible editing property allowing us to control the strength of edit at inference.

\topic{SOTA comparison.}
In Table~\ref{tab:sota_v2v_tgve} we compare several state-of-the-art methods with ours in terms of quality and computation cost. These methods include: Fairy~\cite{wu2024fairy}, a anchor-based cross-frame attention method for fast and coherent video-to-video synthesis; TokenFlow~\cite{geyer2023tokenflow}, a video editing method that propagate inversion features across frames; Rerender-A-Video~\cite{yang2023rerender}, a zero-shot text-guided model with temporal-aware patch matching; ControlVideo~\cite{zhang2023controlvideo}, a training-free controllable model with cross-frame interaction; InsV2V~\cite{insv2v}, a consistent video editing model, EVE~\cite{fdd}, a frame editing method with video adapters, and RAVE~\cite{kara2024rave} that uses a novel noise shuffling strategy for fast zero-shot video editing.
We found these methods to be highly computationally intensive in terms of FLOPs, making them unsuitable for mobile video editing applications. 
In contrast, our series of optimizations demonstrates significant improvements in both TFLOPs and GPU latency per frame while maintaining good quality edits, as evidenced by PickScore~\cite{kirstain2023pickapicopendatasetuser} and CLIPFrame~\cite{hessel2022clipscorereferencefreeevaluationmetric} metrics in the TGVE benchmark~\cite{wu2023cvpr}. Finally, we profile the latency of our method on a Xiaomi 14 Pro with Snapdragon 8 Gen. 3 mobile platform. As shown in Table~\ref{tab:sota_v2v_tgve}, our series of optimizations lead to the on-device denoising pipeline with the impressive frame rate of 12. Check~\cref{sec:appendix:results} for the human evaluation study.
\begin{figure*}[tb]
\centering
\includegraphics[width=2\columnwidth]{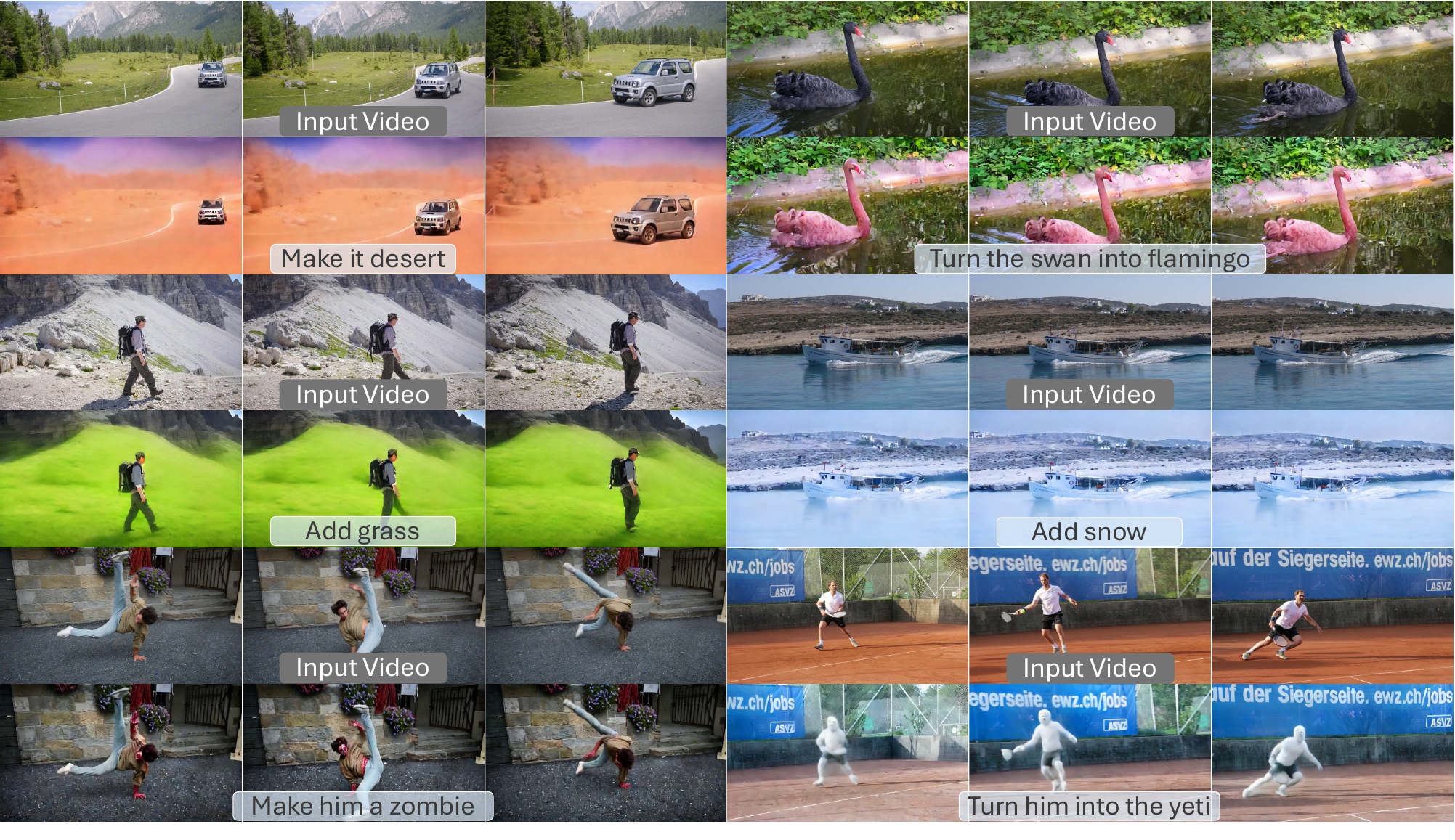}
\caption{Qualitative results of \methodname on DAVIS. Our method can handle complex global edits as well as perform more nuanced attribute editing while requiring very few computational resources. Please refer to the Appendix for video results.}
\label{fig:qual_results_davis}
\end{figure*}

\begin{figure}[tb]
\centering
\includegraphics[width=1\columnwidth]{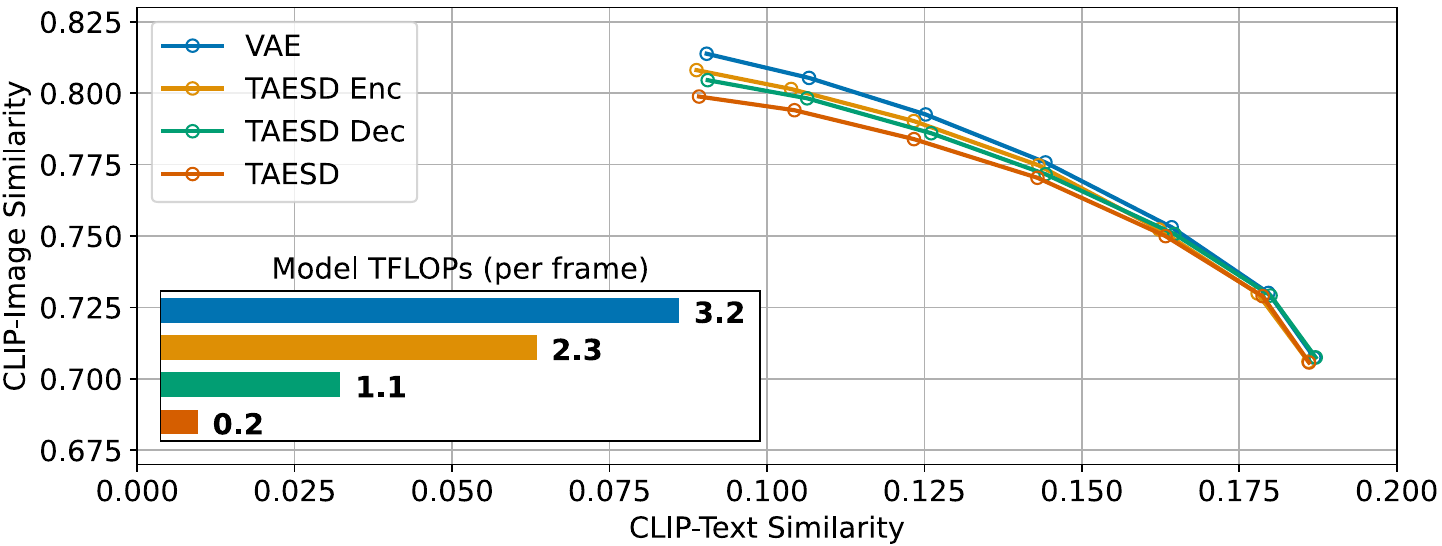}
\caption{CLIP metrics for different autoencoder configurations. A substational FLOPs reduction can be achieved by incorporating TAESD, with minimal drop in editing performance. 
}
\label{fig:vae_ablations_clip_plot}
\end{figure}

\begin{figure*}[tb]
\centering
\includegraphics[width=2.01\columnwidth]{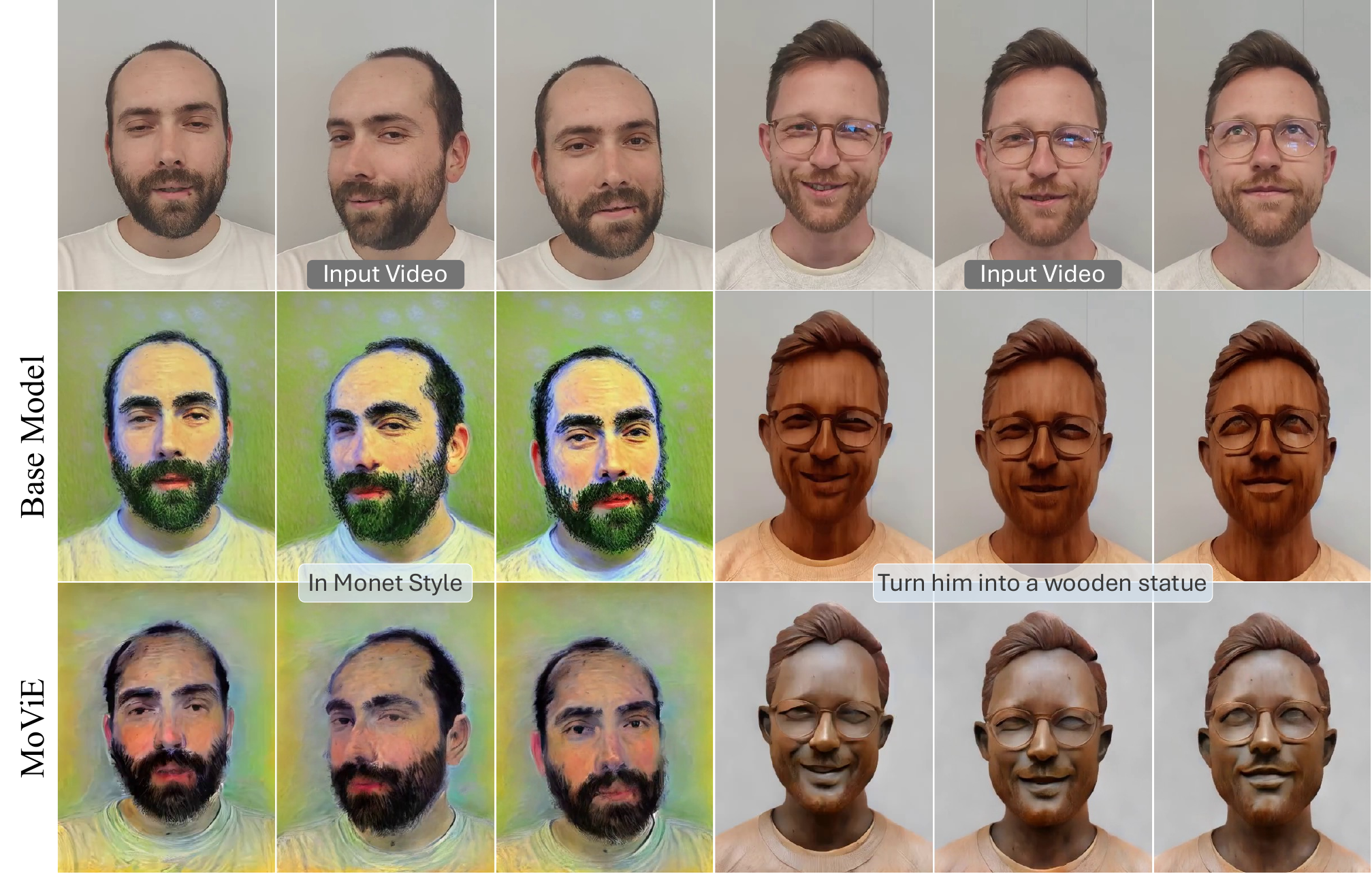}
\caption{Qualitative comparison of our method to the base model. The efficiency is greatly improved whereas quality is not compromised both for style transfer and attribute edits. Please refer to the Appendix for video results.}
\label{fig:qual_results_faces}
\end{figure*}

\begin{figure*}[tb]
\centering
\includegraphics[width=2.01\columnwidth]{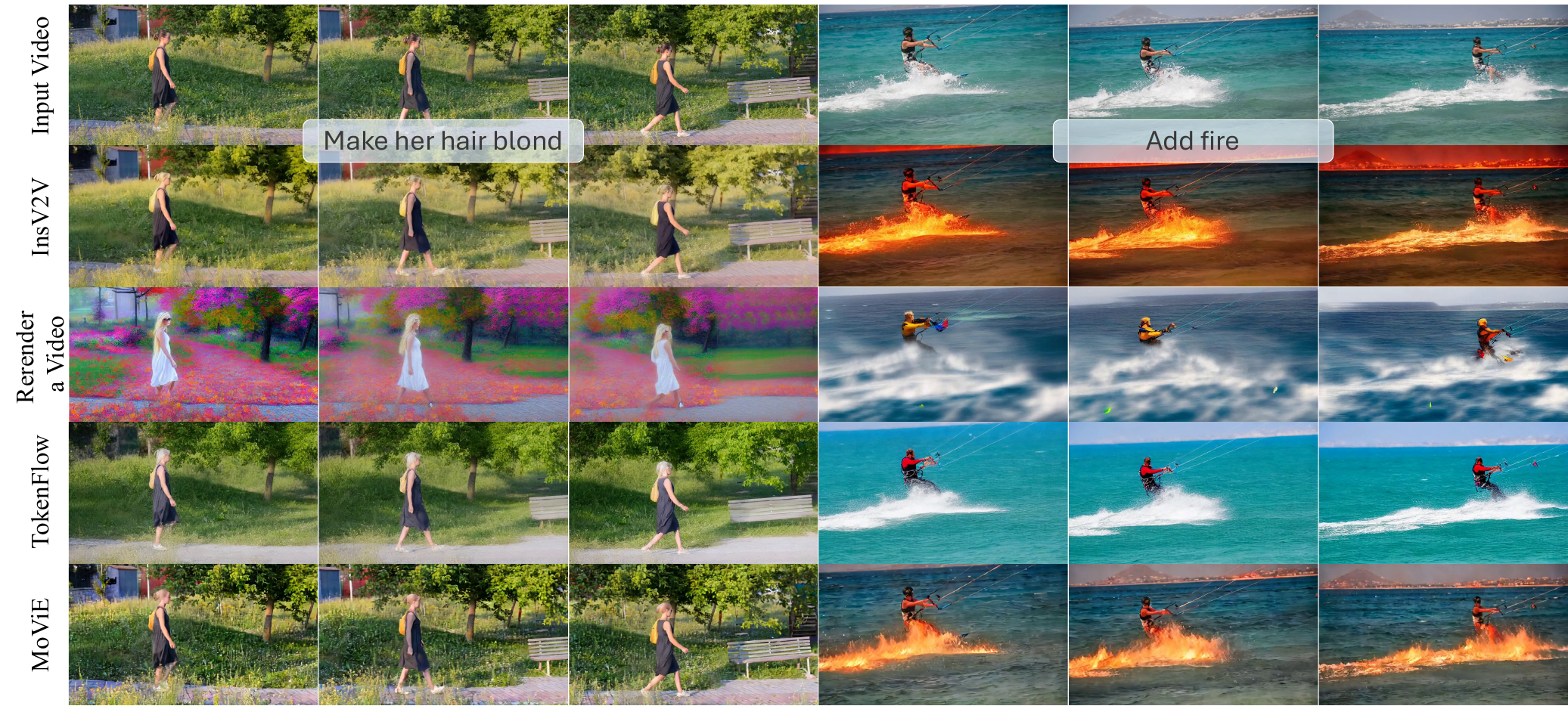}
\caption{Qualitative comparison of our method \methodname to other SOTA video editing algorithms. We evaluate on two challenging editing scenarios. \methodname produces good quality edits yet far outperforms other competing methods in terms of efficiency. Please refer to the Appendix for video results.}
\label{fig:qual_results_sota}
\end{figure*}

\topic{Qualitative results.}
First, we present the qualitative results of \methodname on DAVIS~\cite{davisdataset} videos, as illustrated in Figure~\ref{fig:qual_results_davis}. Our method effectively performs both global edits and fine-grained attribute editing with high efficiency. In Figure~\ref{fig:qual_results_faces}, we compare our model to the base model, demonstrating enhanced efficiency while maintaining editing quality. Finally, in Figure~\ref{fig:qual_results_sota}, we compare \methodname with several state-of-the-art methods on two DAVIS~\cite{davisdataset} videos featuring challenging editing prompts. In the first scenario, which involves changing the hair color of a woman, InsV2V~\cite{insv2v} and TokenFlow~\cite{geyer2023tokenflow} performed relatively well, whereas Rerender-A-Video~\cite{yang2023rerender} introduced numerous unnecessary changes. In the second scenario, where a fire effect was requested to be added in, most methods failed except for InsV2V~\cite{insv2v}. For both prompts, our method produced good quality edits, indicating its competitiveness with more expensive video editing algorithms. Please refer to ~\url{https://qualcomm-ai-research.github.io/mobile-video-editing/} for video results.

\subsection{Ablations}
\label{sec:ablations}
\topic{Adversarial Training.}
\begin{table}[t!]
\centering
\resizebox{\columnwidth}{!}{
\def\arraystretch{1}%
\begin{tabular}{lcccccccc}
\toprule
\centering
   \textbf{Type of Ablation } & \textbf{Ablation detail} & \textbf{CLIP-Image} \\
    \midrule
    \multirow{3}{*}{Noise Distribution}     & $m=0$,  $s=1$ & 0.759 \\
                                            & $m=-1$, $s=1$ & 0.769 \\
                                            & $m=-1$, $s=2$ & 0.765 \\
    \midrule
    \multirow{2}{*}{Head Architecture}      & No guidance conditioning & 0.781 \\
                                            & With guidance conditioning & 0.786 \\
\bottomrule
\end{tabular}
}
\caption{Ablation study on discriminator design choices in adversarial training. 
}
\label{tab:discriminator_ablations}
\vspace{-0.7em}
\end{table}

To confirm the effectiveness of our design choices, we conduct an ablation study for both the discriminator and generator setup. We use the guidance-distilled Mobile-Pix2Pix model from \textbf{Stage-3} that has been finetuned for $v-prediction$. We evaluate the models for a single sampling step using the InstructPix2Pix validation dataset and report CLIP-Image similarity metric in~\cref{tab:discriminator_ablations} for guidance scale of 7.5 and image guidance scale of 1.2.
For the discriminator noise (see ~\cref{alg:adv_dist}), we find similar observations as \cite{sauer2024fast, zhang2024sf}, where the noise distribution affects the edit quality. As shown, setting the mean and standard deviation of noise distribution to $-1$ and $1$, respectively gives the best results. We also observe that using guidance conditioning on discriminator heads provides better prompt adherence and flexibility although the CLIP-Image scores do not appear to vary significantly. 

\topic{Impact of TAESD.}
Incorporation of TAESD leads to a significant FLOPs reduction which can be seen in~\cref{tab:sota_v2v_tgve} and~\cref{fig:vae_ablations_clip_plot}. More importantly this is accompanied by a moderate drop in relevant CLIP scores. 
In addition, the shared embedding space enables combinations of VAE's and TAESD's encoders - decoders, for incorporating partial gains to the diffusion pipeline. As shown in~\cref{fig:vae_ablations_clip_plot}, assigning more budget to the decoder, although more compute expensive, seems a better option for providing better and more detailed reconstruction. In this work, we use TAESD for both encoding and decoding as we find it a good balance between quality and efficiency.

\section{Conclusion}
In this work, we have explored several optimizations to accelerate diffusion-based video editing. Firstly, we introduced an architectural enhancement of denoising UNet and a lightweight autoencoder, resulting in our Mobile-Pix2Pix model. Secondly, we performed novel multimodal guidance distillation which consolidates classifier-free guidance inference with text and image as input conditionings into one forward pass per diffusion step, achieving a threefold increase in inference speed. Lastly, we optimized the number of diffusion steps to one through an adversarial distillation procedure, while maintaining the controllability of edits. These optimizations enable $12.5$ frames per second video editing on-device, marking a significant milestone towards real-time text-guided video editing on mobile platforms.

{\small
\bibliographystyle{cvpr2023_stylekit/ieee_fullname}
\bibliography{arxiv}
}

\clearpage
\appendix
\maketitlesupplementary
\section{Additional results}
\label{sec:appendix:results}

\subsection{Human Evaluation}
\begin{figure}[tb]
\centering
\includegraphics[ width=\columnwidth]{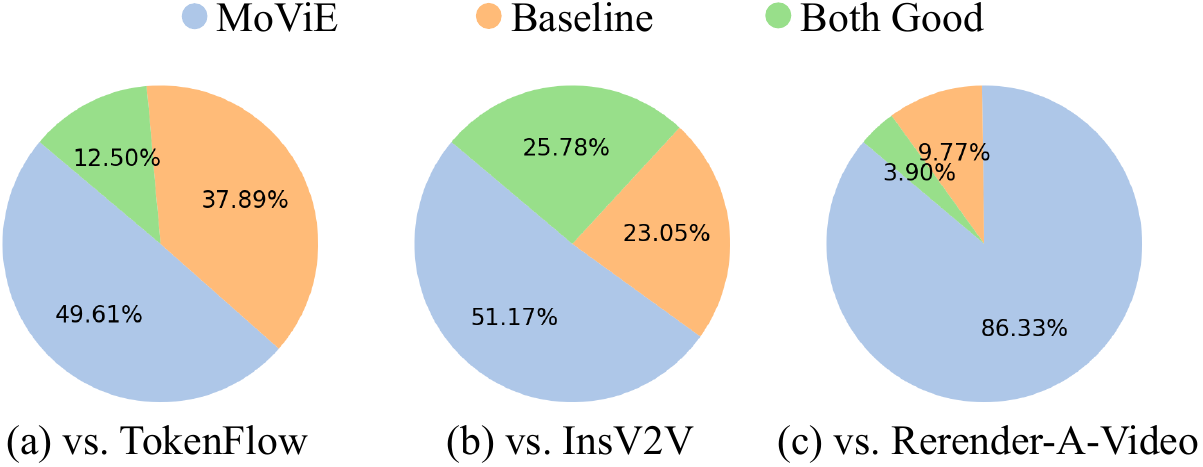}
\caption{Human Evaluation results comparing MoViE to TokenFlow\cite{geyer2023tokenflow}, InsV2V\cite{insv2v}, and Rerender-A-Video\cite{yang2023rerender}.}
\label{sup:fig:human_eval}
\end{figure}

We conducted an A/B comparison between \methodname and three state-of-the-art (SOTA) methods: TokenFlow~\cite{geyer2023tokenflow}, InsV2V~\cite{insv2v}, and Rerender-A-Video~\cite{yang2023rerender}. Participants were asked to compare the overall quality of the generated videos, choosing between: 1) video 1 is better, 2) video 2 is better, or 3) both are equal. Each participant compared 24 video pairs, with each pair evaluated by 32 participants. As shown in \cref{sup:fig:human_eval}, participants generally preferred our method over the SOTA methods. Our model significantly outperformed Rerender-A-Video and was reasonably preferred over TokenFlow and InsV2V. However, due to the limited sample size, we cannot draw definitive conclusions. 

\section{Training Details}
\label{sec:appendix:training_details}

\begin{figure}[tb]
\centering
\includegraphics[width=1.0\columnwidth]{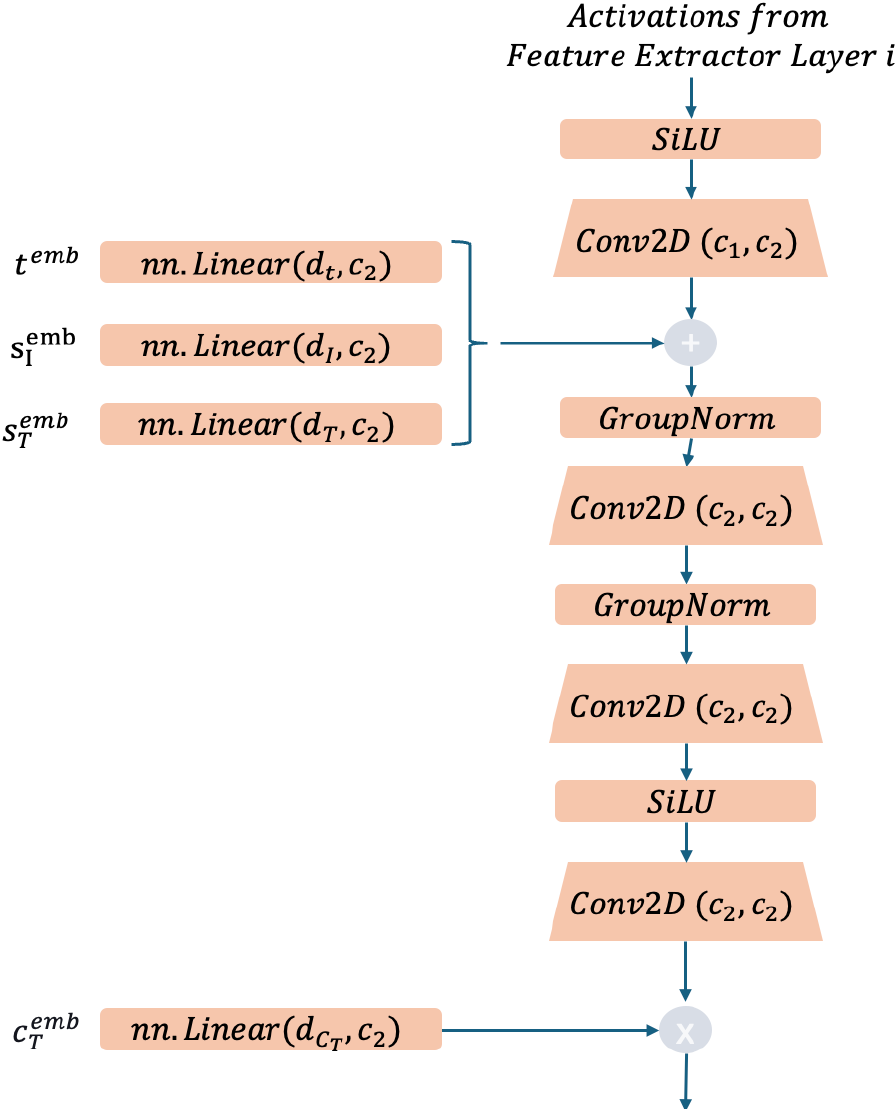}
\caption{Adversarial Head Architecture.}
\label{app:spatial_head}
\end{figure}

\paragraph{Spatial Head Architecture.}
As discussed in Section 3.4, discriminator $D_\phi$ consists of a frozen feature extractor and trainable spatial heads applied to activations received at each layer of feature extractor. In \cref{app:spatial_head} we show a schematic view of the spatial heads. Each spatial head receives as input the activation from layer $i$ of feature extractor along with guidance scale $(s_T^{emb})$, image guidance scale $(s_I^{emb})$ and diffusion timestep $(t^{emb})$ embeddings. We condition the output of each spatial head on the prompt embedding $c_T^{emb}$.

\paragraph{Training.}
Here we provide a detailed description of our training procedure for each stage in our pipeline.
Throughout the finetuning stages we use InstructPix2Pix\cite{ip2p} dataset, batch size of 512 at $256\times256$ resolution and Adam Optimizer \cite{kingma2017adam}.
\par
\textbf{Stage-3.} Our multimodal guidance distillation algorithm is specified in Algorithm 1 of the main manuscript. For training, we use a learning rate of $10^{-5}$. We train on a single A100 GPU and training takes around 14 hours. \par
\textbf{Stage-4.} Image-editing UNet model is an $\epsilon-prediction$ model. However, we find that prior to adversarial training, finetuning the UNet model for $v-prediction$ helped to reduce artifacts during training. We use a teacher-student set-up and train the student to match the $v-prediction$ from multimodal guidance distilled teacher obtained in Stage-3 (We obtain $v-prediction$ from $\epsilon-prediction$ using the standard equation as in \cite{progressivedistill}). For training, we use a learning rate of $10^{-6}$ with 4k warm-up steps. Training takes about a day. \par 

Our adversarial distillation pipeline is shown in Figure 3 and the corresponding training algorithm is specified in Algorithm 2 of the main manuscript. We train for 20k iterations and learning rate of $10^{-5}$. Training takes about 3 days on a single A100 GPU.\par

Here we add further details regarding our training pipeline. 
We provide details of each training component here.
\begin{itemize}
    \item \textbf{Teacher.} Teacher model is initialized from the UNet checkpoint obtained in \textbf{Stage-3} and remains frozen. We evaluate the teacher model for 5 sampling steps to obtain the clean latent $x_0$ using $LCM$ scheduler \cite{luo2023latentconsistencymodelssynthesizing}. We find that the teacher checkpoint generates good quality results for 5 sampling steps.
    \item \textbf{Student.} Student model is  initialized from the UNet checkpoint obtained after the $v-prediction$ finetuning described above. Student is trained to denoise the latent and generate a clean latent $\hat{x_0}$ at each sampled diffusion timestep $t$. During training, we use $Euler Discrete$ \cite{karras2022elucidatingdesignspacediffusionbased} noise scheduler with 8 training timesteps. During evaluation, we evaluate the student for a single timestep using $LCM$ scheduler. 
    \item \textbf{Discriminator.} Discriminator consists of Feature Extractor and spatial heads. We only train the spatial heads while the feature extractor remains frozen.
    Discriminator uses a separate $Euler Discrete$  \cite{karras2022elucidatingdesignspacediffusionbased} noise scheduler with 1000 training timesteps. During training, we find that fixing discriminator noise distribution to logit-normal as in \cite{zhang2024sf} and using $mean=-1,std=1$ gave us the best performance. We ablate this property in Section 4.3.
\end{itemize}

\end{document}